\pdfoutput=1

\documentclass[11pt]{article}

\usepackage[final]{acl}

\usepackage{balance}
\usepackage{times}
\usepackage{soul}
\usepackage{longtable}
\usepackage{latexsym}
\usepackage{graphicx}
\usepackage{color,colortbl}
\usepackage{adjustbox}
\usepackage{array}
\usepackage{amsmath}
\usepackage{caption}
\usepackage{booktabs}
\usepackage{multirow}
\usepackage{microtype}
\usepackage[utf8]{inputenc}
\usepackage[T1]{fontenc}
\usepackage{xcolor}
\usepackage{float} 
\usepackage{multicol}
\usepackage[normalem]{ulem}
\usepackage{flafter}
\definecolor{lightgray}{gray}{0.9}

\newcolumntype{R}[2]{%
    >{\adjustbox{angle=#1,lap=\width-(#2)}\bgroup}%
    l%
    <{\egroup}%
}

\newcommand{\abg}{\textcolor{orange}}

\usepackage[T1]{fontenc}

\usepackage[utf8]{inputenc}

\usepackage{microtype}

\definecolor{Gray}{gray}{0.9}
\definecolor{MGray}{gray}{0.75}
\definecolor{LGray}{gray}{0.5}

\title{Anna Karenina Strikes Again: \\ Pre-Trained LLM Embeddings May Favor High-Performing Learners}
\author{Abigail Gurin Schleifer$^1$ \ \  Beata Beigman Klebanov$^2$ \ \ Moriah Ariely$^1$ \ \   Giora Alexandron$^1$\\
  $^1$ Weizmann Institute of Science, Rehovot, Israel \\
  $^2$ ETS, Princeton, USA \\
  \texttt{\{abigail.gurin-schleifer,moriah.ariely,giora.alexandron\}}\\
  \texttt{@weizmann.ac.il}\\
  \texttt{bbeigmanklebanov@ets.org}\\
}

\begin{document}
\maketitle       
\begin{abstract}
Unsupervised clustering of student responses to open-ended questions into behavioral and cognitive profiles using pre-trained LLM embeddings is an emerging technique, but little is known about how well this captures pedagogically meaningful information. We investigate this in the context of student responses to open-ended questions in biology, which were previously analyzed and clustered by experts into theory-driven Knowledge Profiles (KPs). Comparing these KPs to ones discovered by purely data-driven clustering techniques, we report poor discoverability of most KPs, except for the ones including the correct answers. We trace this `discoverability bias' to the representations of KPs in the pre-trained LLM embeddings space. 
\end{abstract}
\section{Introduction}
Classifying students into behavioral or cognitive profiles using unsupervised cluster analysis techniques is a common application of machine learning to educational data \cite{le2023review,martin2023exploring,ariely2024causal,rastrollo2020analyzing,bovo2013clustering}. Recently, there has been a growing interest in applying this methodology to textual student responses that are decoded using pre-trained large language models into vectorized embeddings in semantic spaces \cite{martin2023exploring,wulff2022bridging,masala2021extracting}. The operational appeal of this approach is that it minimizes the need for expert knowledge, which is costly to inject through human labeling procedures \cite{nehm2012human,tansomboon2017designing,li2023can,ariely2024causal}. However, the validity of patterns discovered this way depends on the ability of the embeddings to maintain the pedagogically meaningful information that existed in the original, textual representations of responses \cite{devlin2018bert,seker2022alephbert} and of the algorithmic method to discover them. Evaluation of emergent profiles is often done in terms of the internal quality of the clustering, as data is usually not available to estimate the extent to which the discovered profiles align with a pedagogically meaningful representation of the responses. Without such an evaluation, a loss of important information can be overlooked, 
potentially leading to sub-optimal educational decisions that rely on this analysis \cite{le2023review}.

To investigate whether this hypothesized risk manifests in real-life educational context, we utilize student answers to two constructed response questions in high school biology. The data was previously analyzed by a team of biology education researchers and experienced teachers, and graded according to a theory-driven detailed analytic rubric that is based upon the Causal-Mechanical Explanation framework \cite{ariely2024causal,salmon2006four}. The rubric contained 10 (item 1) or 11 (item 2) binary categories, each checking for the occurrence of a specific key piece of information in the response. Using these human-generated binary vectors of length 10 (11), the responses were clustered using a KMeans algorithm into a set of 6 (7) Knowledge Profiles ({\bf KPs}) that were found by teachers to encapsulate specific patterns of errors. 

The validity of the KPs was evaluated in several ways. First, human experts conducted a qualitative analysis to assess whether each KP captures a specific and distinct pattern of errors. Second, we analyzed the results computationally, showing that i) the KPs were consistent across the two items, namely, revealing the same type of conceptual errors; and ii) the \textit{learners} tended to exhibit the same type of conceptual error (KP) in both items.  Third, we conducted an in-class formative assessment intervention study that provided automated guidance to students based on their KP,  and showed significant improvement in their performance on a different prompt that measures the same conceptual knowledge. These analyses provided strong evidence that the KPs capture pedagogically meaningful information (for full details, see \citet{ariely2022personalized, ariely2024causal}).

Using these data, we are in a position to answer two research questions: 

\begin{description}
\item[RQ1] What is the correspondence between clusters that are computed from pre-trained LLM embeddings of student responses and theory-based KPs?
\end{description}

 To preview the result, we find that two clustering techniques that are commonly used for such tasks \cite{le2023review} -- KMeans \cite{lloyd1982least} and HDBSCAN \cite{McInnes2017} -- largely fail to discover the KPs though retrieval is somewhat better for the profile containing the correct responses. Following up on this finding, we go `upstream', to the pre-trained embeddings, and investigate:
\begin{description}
\item[RQ2]  How well are the KPs represented in the pre-trained embeddings space?
\end{description}

Our results reveal a strong relationship between the quality of the responses in the profile (correct or various degrees of incorrect) and the shape and density of its embeddings-based representation.  We refer to this relationship as an `Anna Karenina principle' and tie it to the profile discovery failure we observed in RQ1.

The contribution of this work is twofold. First, it is the first to demonstrate the Anna Karenina principle in the context of pre-trained representation of student responses to open-ended questions. Second, our results suggest that, in some cases, out-of-the-box pre-trained LLM embeddings may be a pedagogically unsound basis for profile discovery.

\section{Related Work}

\subsection{NLP-based profiling of constructed responses in science education}
Open-ended items require students to develop and construct their answers, reflect on their knowledge, and integrate it with new ideas \cite{fellows1994window}. Reasoning and evidence-based defense of an argument is key for testing scientific hypotheses \cite{toulmin2003uses}.  Therefore, constructing causal explanations is an essential skill for students of science  to learn \cite{ariely2024causal,martin2023exploring}; practice and high-quality feedback are key elements in helping students master the skill \cite{hattie2007power,gerard2016using,tansomboon2017designing}.


Analyzing open-ended items to provide feedback is a time-consuming, complex task. Automating some of the analyses for assessment and feedback purposes is promising for supporting teaching and learning \cite{tansomboon2017designing,gerard2016using,ariely2023machine}.

Most systems for automated evaluation of scientific explanations to date had been designed in the supervised machine learning framework \cite{schleifer2023transformer,sung2019improving,riordan-etal-2020-empirical,kumar2019get,mizumoto-etal-2019-analytic,li-etal-2021-semantic}. Among the unsupervised approaches, \citet{masala2021extracting} extracted the main takeaways from students' feedback on different components in academic courses, using KMeans to cluster pre-trained BERT embeddings of students' feedback. \citet{martin2023exploring} applied HDBSCAN over pre-trained LLM embeddings and to find emergent argumentation patterns' characteristics. \citet{wulff2022bridging} investigated HDBSCAN clustering over LLM embeddings to evaluate the attention of preservice physics teachers to classroom events elicited from open-ended text responses. 
A semi-supervised coding method in which homogeneous clusters receive the same coding automatically and heterogeneous clusters are fully labeled by humans was proposed by \citet{andersen2023semi} and applied to student responses to PISA items.

\subsection{Biases in pre-trained LLMs}
While LLMs are powerful meaning representations that undergird the state-of-art systems on a wide range of NLP tasks, they are also known to exhibit a plethora of social biases that could lead to social harm when the models are used in downstream tasks \cite{bender2021dangers}. In a recent review of the current state of research on LLM bias evaluation, \citet{goldfarb-tarrant-etal-2023-prompt} criticize the field for focusing heavily on the upstream, pre-trained LLMs, in most cases without considering the connection to a specific task the LLMs is being put to (68\% of the reports reviewed), citing this as a threat to the predictive validity of bias measurements. 


In fact, the literature that does consider the connection between upstream (intrinsic) and downstream (extrinsic) behavior suggests that it is not straightforward.  Considering static embeddings (e.g., word2vec) and a commonly used bias test, the Word Embedding Association Test (WEAT) \cite{caliskan2017semantics}, Goldfarb-Tarrant and colleagues \cite{goldfarb-tarrant-etal-2021-intrinsic} found no clear relationship with performance of models using the embeddings, as measured by differences in precision and recall of retrieval of the target construct on data from privileged and non-privileged groups. Extension to contextual embeddings and a wider range of tasks and measures yielded similar results \cite{cao-etal-2022-intrinsic,kaneko-etal-2022-debiasing}. Our contribution extends the discussion towards social constructs beyond the typically considered demographic attributes such as gender, race, ethnicity, age towards a distinction that is particularly relevant when dealing with learner data -- that of learners at the more or less advanced state of understanding of the phenomenon under consideration. We are not aware of prior work comparing LLM representations based on knowledge-related profiles; the closest finding in the literature are examples of poorer performance of LLM-based systems on data produced by English language learners with respect to native speakers of English \cite{baffour-etal-2023-analyzing}.
Additionally, we explore LLMs in a relatively low-resource language (Hebrew) in contrast to the bulk of current work that focuses on English or other high-resource languages: In the 90 LLM bias studies evaluated by  \citet{goldfarb-tarrant-etal-2023-prompt}, only two report results in a language that is not highly resourced.

\section{Data} 
\label{sec:data}
The data consists of 669 student responses to two open-ended items in high-school biology, collected anonymously from students in grades 10-12 from about 25 high schools of varied demographics and socioeconomic status (based on location) across Israel. Gender distribution was 70$\%$ females (typical to the gender distribution among high-school biology majors in Israel). The items deal with the connection between respiration and energy in physical activity in the context of smoking ({\bf Q1}) and anemia ({\bf Q2}), taught as part of the core topic ``The human body''. The items were human-scored using a similar analytic rubric containing 10 (Q1) or 11 (Q2) categories \cite{ariely2024causal}. All rubric categories are binary, each targeting specific information that needs to be mentioned in a correct response, such as ``the role of hemoglobin in oxygen transportation''  or ``changes in cellular respiration rate''. The resulting binary vectors were clustered using KMeans; the clusters were analyzed by experienced teachers and ranked from 1 to 6 (Q1) or 7 (Q2) with larger numbers corresponding to clusters with more severe errors. We denote these clusters \textit{Knowledge Profiles}, and index them from 1 (KP1) to 7 (KP7). See \citet{ariely2024causal} for a full description of the items and the assessment framework. The items and examples of student responses and their mapping into KPs can be found in Appendix 1. 

\setcounter{footnote}{0}
For the purposes of the analysis presented in this paper, all responses were represented using rich contextualized vectors -- embeddings produced by a pre-trained Large Language Model (LLM). The LLM being used, AlephBERT \cite{seker2022alephbert}, is state-of-the-art for Hebrew. It was trained on a large corpus of the Hebrew language, including: Twitter tweets, Hebrew Wikipedia, and the Hebrew subset of the  Oscar \cite{suarez2020monolingual} dataset. 
AlephBERT has the same architecture as BERT \cite{devlin2018bert}: 12 layers, 110M parameters, and 12 attention heads. It was trained on a  52K-word Hebrew vocabulary on masked token prediction task, and on the Hebrew language tasks: word segmentation, part-of-speech tagging, and full morphological tagging. It was further trained on the tasks of sentiment analysis and named entity recognition.

\section{Methods}
\label{sec:methods}
To evaluate whether raw LLM embeddings carry useful knowledge for unsupervised profiling of responses, we experimented with two common clustering approaches \cite{le2023review}, KMeans \cite{lloyd1982least} and HDBSCAN \cite{McInnes2017}, which implement different clustering mechanisms. The first discovers convex-shaped clusters; its mechanism is centroid-based and applies an Euclidean distance function. The second is density-based and can be applied with various distance metrics, e.g., a metric induced by cosine-similarity, and the clusters may have various shapes. Both approaches were used previously for profile discovery in constructed response data 
\cite{ariely2024causal,martin2023exploring,wulff2022bridging}.
Experiments were conducted in Python, using scikit-learn \cite{scikit-learn}, SBERT \cite{reimers2019sentence} and Pytorch \cite{paszke2019pytorch}. 

\subsection{KMeans}
The KMeans is a widely used algorithm \cite{lloyd1982least}. 
The algorithm is initiated with a specified number of clusters and a random initialization of their centroids. The clustering approach minimizes the within-cluster sum of squared distances, i.e., Euclidean distance.  KMeans clusters are convex and all samples are assigned to a cluster, i.e., there are no outliers. 
Convexity means that for every two points in the cluster, a straight line between them also lies within the cluster.




\subsection{Hierarchical Density-Based Spatial Clustering of Applications with Noise (HDBSCAN)}
Another clustering approach, which is more promising in the context of LLMs' embeddings \cite{martin2023exploring} is the HDBSCAN \cite{McInnes2017} algorithm. The approach here is creating a mutual reachability graph where  \textit{core} samples are points in areas of high density. A cluster is a set of \textit{core} samples and a set of \textit{non-core} samples that are neighbors of \textit{core} samples but are not \textit{core}  themselves. \textit{Non-core} samples are at the fringes of clusters. A \textit{core} sample is such that there are 'min\_samples' other samples with a distance less than $\epsilon$ from it, for some $\epsilon > 0$ \cite{scikit-learn}. The HDBSCAN mechanism performs clustering for various $\epsilon$ values and the most stable clustering is chosen. 



The default metric for HDBSCAN is Euclidean distance. To use cosine similarity, we turn it into a distance function \cite{McInnes2017}: 
\begin{equation}
    \|x-y\|=\sqrt{2\times(1-CosSim(x,y))}\ \label{eq: 1},    
\end{equation}
where $x,y$ are unit vectors, i.e., $\|x\|=\|y\|=1$ \cite{Manning1999found}. 
Since cosine similarity does not depend on vectors' magnitude, only on the angle between the two vectors, we first turned every embedding $e_i$ to a unit vector $\frac{e_i}{\|e_i\|}$ and then applied the HDBSCAN on a pre-computed metric matrix consisting of all pairwise distances between all embeddings in the dataset using formula (\ref{eq: 1}).

In contrast to the KMeans,  HDBSCAN can find clusters with varied densities and clusters may have non-convex shapes.  

\subsection{Metrics for Comparing Clusters}
To compare the similarity between the KPs and the cluster assignments of the KMeans/HDBSCAN, we used 
Adjusted Rand Index (ARI) \cite{vinh2009information}. In ARI, similarity is interpreted as the number of pairs of items on which the clusterings agree,
adjusted for the amount of chance agreement. Let $D$ be a dataset containing $n$ items that are classified into $m$ clusters by clustering C and, independently, into $k$ clusters by clustering E. For a pair of items $(i_1, i_2)\in D$, C and E agree on it iff $i_1$ and $i_2$ are either (1) assigned to the same cluster in both C and E (let's say there are $a$ such pairs), or (2) assigned to different clusters in both C and E (let's say there are $b$ such pairs).  
Now, $a+b$ is the number of agreements between C and E. The ARI index is given by:
\[
RI = \frac{a+b}{\binom{n}{2}}\quad
~;~
\quad ARI = \frac{RI-E[RI]}{\max(RI) - E[RI]}\quad,
\]

where $E[RI]$ is the expected RI for some random label assignment \cite{vinh2009information}, and $max(RI)$ equals to $1$.
The ARI values range from $-1$ to $1$, where $1$ indicates perfect agreement, and $-1$ indicates complete disagreement \cite{hubert1985comparing}.
Since each student response in our dataset is labeled with its KP, we evaluated the ARI for each clustering assignment, i.e., KMeans and HDBSCAN, compared to the KPs. This yields a \textit{global} comparison between the KPs and each clustering assignment.  

To evaluate the `discoverability' of each KP, we also conducted {\bf \textit{by-KP analysis}}, applying a retrieval paradigm and considering each cluster as an attempt to retrieve each of the KPs. We calculate \textit{recall}, \textit{precision}, and \textit{F1} score using a contingency matrix $A=(a_{mn})_{1\leq m\leq k,\ 1\leq n\leq f}$ where  rows are the KPs $k=6,7$, and columns are the unsupervised clusters $C_n$ found by KMeans or HDBSCAN, $f=\#fitted\_clusters$; 
\[
a_{mn}=\big|\{x:\ x\in KP_m\cap C_n\}\big|
\]
the cell $a_{mn}$ in the matrix $A$ counts the number of members of KP$_m$ that fell in cluster C$_n$.  The precision of retrieval of KP$_m$  using cluster C$_n$  is $P_{mn}=\frac{a_{mn}}{|C(n)|}$;  the recall is  $R_{mn}=\frac{a_{mn}}{|KP(m)|}$. F1 score is  $F_{mn}=\frac{2\cdot P_{mn}\cdot R_{mn}}{P_{mn}+R_{mn}}$, indicating the extent to which we were able to retrieve KP$_m$ using the emergent cluster C$_n$. 



\section{Results}
\label{results}
\subsection{RQ1: Correspondence between embedding-based clusters and theory-based Knowledge Profiles
}



\subsubsection{Global alignment between the clusterings} 
As described in Section~\ref{sec:methods}, we evaluated the agreement between clusterings that were computed from the embeddings using two cluster analysis methods: KMeans and HDBSCAN. As we were interested in upper-bounding the discoverability of the KPs by both algorithms, we ``helped" them with additional information (the number of clusters to the KMeans algorithm, and allowing the HDBSCAN to grid search for `good' hyperparameters).   With $k$  equals the number of KPs per item (six for Q1 and seven for Q2), the resulting ARIs for the KMeans were \textbf{0.122} and \textbf{0.191}, for Q1 and Q2, respectively. For the HDBSCAN algorithm, we conducted a grid search for two parameters: $min\_cluster\_size$, i.e., the minimum number of samples in a cluster (values:$\{3,4,5,10,15,20,30,40\}$), and $min\_samples$, i.e., the number of samples in a neighborhood for a point to be considered as a core point (values: $\{1,2,3,4,5\}$). We report the best-performing combination in terms of ARI:  
$\textbf{0.037}$ for Q1 (with $min\_cluster=5$, $min\_samples=2$), and  $\textbf{0.038}$ for Q2 (with $min\_cluster=3$, $min\_samples=3$).
Based on these results, we conclude that the  clusters discovered by the KMeans had low agreement with the KPs, and the clusters discovered by the HDBSCAN had negligible agreement with the KPs.

\subsubsection{Discoverability of specific KPs}
We further investigated the clusters' matching quality by calculating the F1 score \textit{per KP} for each of Q1 and Q2. For KMeans, the results show good retrieval of KP1, the cluster with the highest-quality responses -- $F1=0.60,\ 0.67$ for items Q1 and Q2 respectively -- but much worse retrieval of the other KPs, with maximal $F1=0.40$ for KP6 in Q1 and $F1=0.47$ for KP2 in Q2. The clustering results in terms of contingency tables and  F1 Scores are presented in Tables~\ref{tab:Q1_KM} to~\ref{tab:Q2_KM2}, with KPs as rows and columns as fitted clusters.The maximum F1 scores per profile are shaded in gray.



\begin{table*}[t]
    \begin{minipage}[t]{.45\linewidth}
        \begin{minipage}{\linewidth}
            \centering
            \begin{tabular}{|c|>{\centering\arraybackslash}p{0.55cm}|>{\centering\arraybackslash}p{0.55cm}|>{\centering\arraybackslash}p{0.55cm}|>{\centering\arraybackslash}p{0.55cm}|>{\centering\arraybackslash}p{0.55cm}|>{\centering\arraybackslash}p{0.55cm}|}
                \hline
                 &  \textbf{A} & \textbf{B} & \textbf{C} & \textbf{D} & \textbf{E} & \textbf{F}\\
                \hline
                \textbf{KP1} & 0 & 102 & 0 & 0 & 5 & 24\\
                \textbf{KP2} & 3 & 39 & 0 & 0 & 12 & 37\\
                \textbf{KP3} & 3 & 22 & 0 & 0 & 13 & 65\\
                \textbf{KP4} & 13 & 28 & 0 & 0 & 26 & 39\\
                \textbf{KP5} & 12 & 14 & 0 & 0 & 36 & 50\\
                \textbf{KP6} & 31 & 5 & 3 & 16 & 55 & 16\\
                \hline
            \end{tabular}
            \caption{Q1 KMeans contingency matrix.}\label{tab:Q1_KM}
        \end{minipage}
    \end{minipage}%
    \hspace{0.7cm}
    \begin{minipage}[t]{.55\linewidth}
        \begin{minipage}{0.88\linewidth}
            \centering
            \begin{tabular}{|c|>{\centering\arraybackslash}p{0.65cm}|>{\centering\arraybackslash}p{0.65cm}|>{\centering\arraybackslash}p{0.65cm}|>{\centering\arraybackslash}p{0.6cm}|>{\centering\arraybackslash}p{0.65cm}|>{\centering\arraybackslash}p{0.6cm}|}
                \hline
                 &  \textbf{A} & \textbf{B} & \textbf{C} & \textbf{D} & \textbf{E} & \textbf{F}\\
                \hline
                \textbf{KP1} & .00 & \cellcolor{lightgray}.60 & .00 & .00 & .04 & .13\\
                \textbf{KP2} & .04 & \cellcolor{lightgray}.26 & .00 & .00 & .10 & .23\\
                \textbf{KP3} & .04 & .14 & .00 & .00 & .10 & \cellcolor{lightgray}.39\\
                \textbf{KP4} & .15 & .18 & .00 & .00 & .21 & \cellcolor{lightgray}.23\\
                \textbf{KP5} & .14 & .09 & .00 & .00 & .28 & \cellcolor{lightgray}.29\\
                \textbf{KP6} & .33 & .03 & .05 & .23 & \cellcolor{lightgray}.40 & .09\\
                \hline
            \end{tabular}
            \caption{Q1 KMeans  F1 Scores.}\label{tab:Q1_KM2}
        \end{minipage}
    \end{minipage}
\end{table*}

\begin{table*}[t]
    \begin{minipage}[t]{.45\linewidth}
        \begin{minipage}{\linewidth}
            \centering
            \begin{tabular}{|c|>{\centering\arraybackslash}p{0.4cm}|>{\centering\arraybackslash}p{0.4cm}|>{\centering\arraybackslash}p{0.4cm}|>{\centering\arraybackslash}p{0.4cm}|>{\centering\arraybackslash}p{0.4cm}|>{\centering\arraybackslash}p{0.4cm}|>{\centering\arraybackslash}p{0.4cm}|}
                \hline
                 &  \textbf{A} & \textbf{B} & \textbf{C} & \textbf{D} & \textbf{E} & \textbf{F} & \textbf{G}\\
                \hline
                \textbf{KP1} & 10 & 0 & 2 & 13 & 13 & 127 & 0\\
                \textbf{KP2} & 51 & 0 & 22 & 13 & 4 & 30 & 0\\
                \textbf{KP3} & 11 & 0 & 10 & 21 & 14 & 28 & 0\\
                \textbf{KP4} & 12 & 0 & 18 & 25 & 16 & 13 & 5\\
                \textbf{KP5} & 5 & 0 & 16 & 23 & 29 & 6 & 1\\
                \textbf{KP6} & 3 & 0 & 29 & 8 & 8 & 6 & 10\\
                \textbf{KP7} & 4 & 9 & 20 & 11 & 8 & 3 & 12\\
                \hline
            \end{tabular}
            \caption{Q2 KMeans contingency matrix.}\label{tab:Q2_KM}
        \end{minipage}
    \end{minipage}%
    \hspace{0.6cm}
    \begin{minipage}[t]{.55\linewidth}
        \begin{minipage}{0.88\linewidth}
            \centering
            \begin{tabular}{|c|>{\centering\arraybackslash}p{0.5cm}|>{\centering\arraybackslash}p{0.5cm}|>{\centering\arraybackslash}p{0.5cm}|>{\centering\arraybackslash}p{0.5cm}|>{\centering\arraybackslash}p{0.5cm}|>{\centering\arraybackslash}p{0.5cm}|>{\centering\arraybackslash}p{0.5cm}|}
                \hline
                 &  \textbf{A} & \textbf{B} & \textbf{C} & \textbf{D} & \textbf{E} & \textbf{F} & \textbf{G}\\
                \hline
                \textbf{KP1} & .08 & 0 & .01 & .09 & .10 & \cellcolor{lightgray}.67 & 0\\
                \textbf{KP2} & \cellcolor{lightgray}.47 & 0 & .19 & .11 & .04 & .18 & 0\\
                \textbf{KP3} & .12 & 0 & .10 & \cellcolor{lightgray}.21 & .16 & .19 & 0\\
                \textbf{KP4} & .13 & 0 & .17 & \cellcolor{lightgray}.25 & .18 & .09 & .09\\
                \textbf{KP5} & .06 & 0 & .16 & .24 & \cellcolor{lightgray}.34 & .04 & .02\\
                \textbf{KP6} & .04 & 0 & \cellcolor{lightgray}.32 & .09 & .10 & .04 & .22\\
                \textbf{KP7} & .05 & .24 & .22 & .12 & .10 & .02 & \cellcolor{lightgray}.25\\
                \hline
            \end{tabular}
            \caption{Q2 KMeans F1 Scores.} \label{tab:Q2_KM2}
        \end{minipage}
    \end{minipage}
\end{table*}

The evaluation of HDBSCAN clusters mirrored that of KMeans, showing better retrieval of KP1 --  $0.43, 0.46$ F1 scores for Q1 and Q2 -- than of any other profile, with maximal $F1=0.36$ for KP6 in Q1 and $F1=0.29$ for KP2 in Q2. We observe that, overall, results are worse for HDBSCAN than for KMeans.  The clustering results in terms of contingency tables and  F1 Scores are presented in Tables~\ref{tab:Q1_HDBSCAN} to~\ref{tab:Q2_HDBSCAN2}, with KPs as rows and columns as fitted clusters. The maximum F1 scores are shaded in gray.

\begin{table*}[t]
    \begin{minipage}{.5\linewidth}
        \begin{minipage}{0.95\linewidth}
            \centering
            \begin{tabular}{|c|>{\centering\arraybackslash}p{0.5cm}|>{\centering\arraybackslash}p{0.5cm}|>{\centering\arraybackslash}p{0.5cm}|>{\centering\arraybackslash}p{0.5cm}|>{\centering\arraybackslash}p{0.5cm}|>{\centering\arraybackslash}p{0.5cm}|}
                \hline
                 &  \textbf{A} & \textbf{B} & \textbf{C} & \textbf{D} \\
                \hline
                \textbf{KP1} & 19 & 0 & 112 & 0\\
                \textbf{KP2} & 29 & 0 & 62 & 0\\
                \textbf{KP3} & 36 & 0 & 67 & 0\\
                \textbf{KP4} & 49 & 0 & 57 & 0\\
                \textbf{KP5} & 61 & 0 & 51 & 0\\
                \textbf{KP6} & 71 & 7 & 43 & 5\\
                \hline
            \end{tabular}
            \caption{Q1 HDBSCAN contingency matrix.}\label{tab:Q1_HDBSCAN}
        \end{minipage}
    \end{minipage}%
    \begin{minipage}{.5\linewidth}
        \begin{minipage}{0.95\linewidth}
            \centering
            \begin{tabular}{|c|>{\centering\arraybackslash}p{0.7cm}|>{\centering\arraybackslash}p{0.8cm}|>{\centering\arraybackslash}p{0.8cm}|>{\centering\arraybackslash}p{0.8cm}|>{\centering\arraybackslash}p{0.8cm}|>{\centering\arraybackslash}p{0.8cm}|}
                \hline
                 &  \textbf{A} & \textbf{B} & \textbf{C} & \textbf{D}\\
                \hline
                \textbf{KP1} & .10 & 0 & \cellcolor{lightgray}.43 & 0\\
                \textbf{KP2} & .16 & 0 & \cellcolor{lightgray}.26 & 0\\
                \textbf{KP3} & .20 & 0 & \cellcolor{lightgray}.27 & 0\\
                \textbf{KP4} & \cellcolor{lightgray}.26 & 0 & .23 & 0\\
                \textbf{KP5} & \cellcolor{lightgray}.32 & 0 & .20 & 0\\
                \textbf{KP6} & \cellcolor{lightgray}.36 & .11 & .17 & .08\\
                \hline
            \end{tabular}
            \caption{Q1 HDBSCAN F1 Scores.}\label{tab:Q1_HDBSCAN2}
        \end{minipage}
    \end{minipage}
\end{table*}

\begin{table*}[t]
    \begin{minipage}{.5\linewidth}
        \begin{minipage}{0.95\linewidth}
            \centering
            \begin{tabular}{|c|>{\centering\arraybackslash}p{0.7cm}|>{\centering\arraybackslash}p{0.8cm}|>{\centering\arraybackslash}p{0.8cm}|>{\centering\arraybackslash}p{0.8cm}|}
                \hline
                 &  \textbf{A} & \textbf{B} & \textbf{C} & \textbf{D}\\
                \hline
                \textbf{KP1} & 23 & 0 & 142 & 0\\
                \textbf{KP2} & 39 & 0 & 81 & 0\\
                \textbf{KP3} & 27 & 0 & 57 & 0\\
                \textbf{KP4} & 30 & 0 & 56 & 3\\
                \textbf{KP5} & 32 & 0 & 48 & 0\\
                \textbf{KP6} & 33 & 0 & 31 & 0\\
                \textbf{KP7} & 32 & 3 & 32 & 0\\
                \hline
            \end{tabular}
            \caption{Q2 HDBSCAN contingency matrix.}\label{tab:Q2_HDBSCAN}
        \end{minipage}
    \end{minipage}%
    \begin{minipage}{.5\linewidth}
        \begin{minipage}{0.95\linewidth}
            \centering
            \begin{tabular}{|c|>{\centering\arraybackslash}p{0.7cm}|>{\centering\arraybackslash}p{0.8cm}|>{\centering\arraybackslash}p{0.8cm}|>{\centering\arraybackslash}p{0.8cm}|}
                \hline
                 &  \textbf{A} & \textbf{B} & \textbf{C} & \textbf{D}\\
                \hline
                \textbf{KP1} & .12 & 0 & \cellcolor{lightgray}.46 & 0\\
                \textbf{KP2} & .23 & 0 & \cellcolor{lightgray}.29 & 0\\
                \textbf{KP3} & .18 & 0 & \cellcolor{lightgray}.21 & 0\\
                \textbf{KP4} & .20 & 0 & \cellcolor{lightgray}.21 & .07\\
                \textbf{KP5} & \cellcolor{lightgray}.22 & 0 & .18 & 0\\
                \textbf{KP6} & \cellcolor{lightgray}.24 & 0 & .12 & 0\\
                \textbf{KP7} & \cellcolor{lightgray}.23 & .09 & .12 & 0\\
                \hline
            \end{tabular}
            \caption{Q2 HDBSCAN F1 Scores.} \label{tab:Q2_HDBSCAN2}
        \end{minipage}
    \end{minipage}
\end{table*}

We then considered the possibility that more coarse-grained profiles might emerge from the clustering than the detailed KPs. To this end, we tried different options for grouping KPs and calculated the F1 scores between fitted clusters and the grouped KPs.  The best results show an emergent pattern consistent in both items Q1 and Q2, where one cluster consists of the higher-quality responses (KP1-KP4) an is retrievable with $F1=0.72, 0.74$, and the other cluster consists of lower-quality responses (KP5-KP7), retrievable with $F1=0.52, 0.45$.




We tried this approach with the KMeans as well, but the samples scattering across fitted clusters there did not exhibit meaningful patterns.

\subsection{RQ2: How well are the KPs represented in the embeddings?}
To answer this question, we first analyzed, descriptively, the level of similarity between the embeddings within each KP, and between KPs. We then  conducted statistical tests to verify that the observed patterns are statistically robust.

\noindent\textbf{\textit{Within KP similarity.}} To analyze the level of similarity  within each KP, we computed the pairwise cosine-similarity between all pairs in that KP. Tables~\ref{tab:Q1_cossim} and~\ref{tab:Q2_cossim} show the results, with KPs as rows and fitted clusters as columns. Within-KP similarities are in the diagonals. Since the pairwise cosine similarity values are not normally distributed\footnote{The two-sided Kolmogorov-Smirnov test:  test statistic = 0.5, $p < 0.001$}, we report  medians.  The results show that KP1's embeddings  (highest quality responses) have the highest density; as the quality of a response goes down, so does its similarity to other responses with the same pattern of error. \\
\textbf{\textit{Between-KP similarity.}} As can be further seen in Tables~\ref{tab:Q1_cossim} and~\ref{tab:Q2_cossim}, for both items, for every $i>1$ the embeddings of KP$_i$ responses tend to be more similar to the embeddings of KP1  than to embeddings of their own KP (the bolded values in the first row are the largest in each column). This means that  erroneous responses of various types are more similar to the correct responses than to those with the same pattern of error. 



\begin{table*}[t]
\centering
\begin{tabular}{|c|>{\centering\arraybackslash}p{1cm}|>{\centering\arraybackslash}p{1cm}|>{\centering\arraybackslash}p{1cm}|>{\centering\arraybackslash}p{1cm}|>{\centering\arraybackslash}p{1cm}|>{\centering\arraybackslash}p{1cm}|}
\hline
\textbf{KP} &  \textbf{1} & \textbf{2} & \textbf{3} & \textbf{4} & \textbf{5} & \textbf{6}\\
\hline
1 & \textbf{.920} & \textbf{.910} & \textbf{.900} & \textbf{.899} & \textbf{.883} & \textbf{.852}\\
2 &  & .903 & .896 & .889 & .880 & .849\\
3 &  &  & .897 & .890 & .883 & .852\\
4 &  &  &  & .877 & .871 & .837\\
5 &  &  &  &  & .874 & .845\\
6 &  &  &  &  &  & .755\\
\hline
\end{tabular}
\caption{Q1 pairwise cosine similarity median per KP.}\label{tab:Q1_cossim}
\end{table*}
\begin{table*}[htbp]
\centering
\begin{tabular}{|c|>{\centering\arraybackslash}p{1cm}|>{\centering\arraybackslash}p{1cm}|>{\centering\arraybackslash}p{1cm}|>{\centering\arraybackslash}p{1cm}|>{\centering\arraybackslash}p{1cm}|>{\centering\arraybackslash}p{1cm}|>{\centering\arraybackslash}p{1cm}|}
\hline
\textbf{KP} &  \textbf{1} & \textbf{2} & \textbf{3} & \textbf{4} & \textbf{5} & \textbf{6} & \textbf{7}\\
\hline
1 & \textbf{.916} & \textbf{.891} & \textbf{.896} & \textbf{.886} & \textbf{.873} & \textbf{.857} & \textbf{.837}\\
2 &  & .876 & .870 & .860 & .854 & .840 & .820\\
3 &  &  & .881 & .870 & .862 & .844 & .828\\
4 &  &  &  & .866 & .860 & .843 & .822\\
5 &  &  &  &  & .861 & .843 & .823\\
6 &  &  &  &  &  & .824 & .796\\
7 &  &  &  &  &  &  & .764\\
\hline
\end{tabular}
\caption{Q2 pairwise cosine similarity median per KP.} \label{tab:Q2_cossim}
\end{table*}

\noindent\textbf{\textit{Hypothesis testing.}} Next, we conducted statistical tests to confirm that i) the distribution of the embeddings in each KP are indeed different and that ii) the cosine similarity within each KP is correlated with the responses quality. 

\noindent \textbf{i)} A Kruskal-Wallis H-test confirmed that at least one of the medians for the different KPs is significantly different from the others, for both Q1 and Q2 (Q1: $statistic=338.435, p<0.001$; Q2: $statistic=295.019, p<0.001$). A follow-up Dunn's post-hoc analysis indicated that the within-KPs similarities differ significantly across all KP pairs, for both Q1 and Q2, with $p< 0.001$. This indicates that embeddings of responses from different KPs have different distributions. Moreover, the embeddings of high-quality responses are highly dense, while embeddings of low-quality responses are more scattered in the vector space. 

\noindent \textbf{ii)} To show that the cosine similarities within KPs are significantly correlated with the responses' quality, we calculated for every sample $x\in KP(i)$ its cosine similarity to $KP(i)$ centroid $c(i)$, where $c(i)$ is the average embedding component-wise of the embeddings in $KP(i)$, i.e., 
\[
CosSim(x,c(i))\quad \forall x\in KP(i).
\]
We then calculated the Spearman correlation between all the similarities values $\cup_{i=1}^{k} \{CosSim(x,c(i)): x\in KP(i)\}$ where $k=6,7$ for Q1, Q2 respectively, and the ordinal variable of the KPs' index, where lower index represents higher-quality responses. 
The Spearman correlation coefficient and its p-values are: 
\[
      r_{Q1}=-0.686,~ p<0.001
\] 
\[
      r_{Q2}=-0.633, p<0.001
\] 

indicating  a strong correlation \cite{xiao2016using} between the quality of a `family of responses' (KP) and the within-family similarity.

\section{Discussion and Conclusion} 
Our data consists of  $669$ high-school student responses to two typical constructed response items in high-school biology. The responses were human graded according to an analytic rubric that is based on the Causal-Mechanical explanation framework \cite{ariely2023machine}, transforming each response to a binary vector that encodes the grading according to the rubric categories. Previous work demonstrated that applying cluster analysis (KMeans) to these vectors, which result from a process that applies a theoretical assessment framework to concrete context by human experts, yields stable clusters that reveal pedagogically meaningful knowledge profiles, which were validated in several ways \cite{ariely2024causal}. (For more details, see Section~\ref{sec:data}.) We reasoned that given the successful performance of pre-trained LLMs on a variety of education-related meaning-intensive tasks \cite{schleifer2023transformer,wambsganss2023unraveling,riordan-etal-2020-empirical,sung2019improving}, and previous work that applied this specifically to profile discovery \cite{martin2023exploring,wulff2022bridging}, we want to evaluate whether  unsupervised profile discovery that is not aided by human knowledge works sufficiently well to be applied out-of-the-box. 

The results of RQ1 reveal that two distinct common unsupervised clustering techniques largely failed to discover the `gold' KPs from the pre-trained LLM embeddings. Inasmuch as a weak relationship with the knowledge profiles was exhibited by KMeans clusters (ARI of 0.12-0.19), our retrieval-based analysis per profile showed that KP1, the profile that captures the correct responses, was the most discoverable profile, with F1-scores of 0.60/0.67 (on Q1/Q2) for retrieving members of KP1 using the best-aligned emergent cluster. Thus, had the emergent clusters been used as a basis for feedback, only the correct responses would have received pedagogically cogent feedback, since responses belonging to low-knowledge KPs are all intermixed in the emergent clusters. This phenomenon was consistent across two items -- Q1 and Q2 -- that were analyzed separately.


In an attempt to account for both the failure of overall profile discovery based on pre-trained LLM embeddings and for the bias towards the correct responses exhibited by the emergent clusters, we turned `upstream' to inspect how the KPs are represented by the embeddings.

We found that the lower the knowledge level of the profile, the less similar to each other its members are in the embeddings space. It is this property that we refer to as the {\bf Anna Karenina principle}: Analogously to Tolstoy's observation that happy families are similar to each other whereas each unhappy family is unhappy in its own way, we see that the correct responses are similar to each other, whereas incorrect responses differ more from each other the more incorrect ('unhappy') they are (strong correlation of $r_1=-0.686, r_2=-0.633$ for Q1/Q2). One could say that Tolstoy considered all unhappy families as an undifferentiated mass; presumably, if classified by their specific source of unhappiness (by family therapists, say), profiles would have probably emerged. In our case, the incorrect responses are grouped by teachers according to the type of problem they exhibit; however, within-profile similarities are still lower than those of correct responses and drift further apart the bigger the problems. The lower density of the poor-knowledge profiles may be one reason that inhibits their downstream discovery.

Further analysis suggests that the privileged status of `happy families' (correct responses) extends beyond their higher density. We also found that while an average correct response is most similar to another correct response, an average incorrect response is closer in the embeddings space to a correct response than to a member of its own profile (Tables~\ref{tab:Q1_cossim} and~\ref{tab:Q2_cossim}). That is, in some sense, the correct responses are the center of the universe, whereas the incorrect responses drift around them in non-convex formations. The non-convexity of the lower-knowledge profiles may be another inhibitor of their downstream discoverability. Taken together, the `classic' Anna Karenina property and the strong version that puts the correct responses in the center suggest an explanation for both the overall discovery failure observed downstream and for the bias in favor of correct responses exhibited by the emergent clusters.

Based on our results, pre-trained representations may not lend themselves to making the necessary distinctions to support pedagogical decisions such as providing formative feedback that targets  specific errors in student reasoning. In particular, our results show a case where the representations are not sufficiently nuanced to allow commonly used clustering methods to identify any error-based profiles, only the profile of the correct responses. Since it is the students who gave the incorrect responses who are in most need of targeted formative feedback, the bias in favor of correct responses is especially counter-productive. Thus, our results tell a cautionary tale about using emergent properties of student response data built over pre-trained embeddings  without domain- and task-specific tuning, and without human supervision.

\subsection{Limitations}
It is possible that other clustering approaches could have revealed clusters that are more similar to the `gold' ones. However, given that despite the large difference between KMeans and HDB SCAN's algorithmic approach, they were quite consistent in both demonstrating poor overall agreement and being biased towards discovering the best KP, we believe that reaching results that are qualitatively different from another clustering method is unlikely.
It is also possible that emergent clusters do correspond to an alternative meaningful partition of the responses into groups, but that partition is not what educators see when they analyze student responses.

The AlephBERT model used in this paper is state-of-the-art for Hebrew, but it has a smaller number of parameters compared to the most recent LLMs for English. It is possible that with more advanced LLM technology, the LLM representations of student responses will be more nuanced; we will revisit our analyses with larger Hebrew LLMs when available.

Due to the monolingual nature of our current data, we have experimented with one language only. Work is underway to collect comparable student response data in Arabic.


%
%
%
\section*{Ethics statement}
We acknowledge that the work is conformant with the ACL Code of Ethics. The research and its data collection procedures were approved by the  Institutional Review Board and the Ministry of Education. The instrument was administered to the students as part of the regular instruction of the topic, based on the teachers' decision to use it as part of the teaching routine (the instrument was published in teachers' forums), with teacher and school principal approval that response data will be used for research.

The goal of this research is to better understand the relations between pre-trained LLM-based representations of student responses to open-ended questions in science, and representations of these responses according to theory-driven rubrics applied by human experts.  We study to what extent conceptually/pedagogically similar responses tend to maintain their proximity in the embedding space as well, and the impact of deviations from this property on downstream analysis. What makes this especially relevant to Ethics is our finding that the weaker students are the ones whose responses suffer the most from representation mismatches between the two representation spaces. This limits the ability to automatically cater to these students -- the ones who are in the highest need for personalized guidance --  with formative feedback that matches the gaps in their reasoning.  By identifying and naming this phenomenon (`the Anna Karenina principle' in automated short answer evaluation), we hope to start a discussion on the means to both estimate its prevalence and to address it.

We demonstrate the Anna Karenina principle on two tasks with one pre-trained model. It is possible that results will look different with other tasks and other large language models. There is a potential danger of over-generalization based on our results, whereby large language models, as a species, so to speak, would be thought to suffer from the Anna Karenina principle and their off-the-shelf use would be avoided in learner-focused applications. This, in turn, could hamper development of useful LLM-based applications to support learners. We believe that the best course of action is to continue the study of the principle in order to improve our understanding of what kind of models are likely to exhibit the problem and for what kind of task, as well as how to diagnose and correct it, ideally without recourse to a large human-tagged dataset. In parallel, future ethics-focused research could investigate whether weaker learners should be a protected category in educational applications, akin to demographic categories like race or gender, by investigating evidence of harm differentially wrought on such learners through technology that does not cater sufficiently precisely to their needs.

Data from this research cannot be shared publicly due to privacy regulations, but may be provided for research purposes, along with its analysis code, subject to the necessary approvals.

\section*{Acknowledgments}
This study was partially supported by the Israeli Science Foundation (ISF) grant number 851/22.

\balance
\bibliographystyle{acl_natbib}
\bibliography{bibliography}

\begin{thebibliography}{44}
\expandafter\ifx\csname natexlab\endcsname\relax\def\natexlab#1{#1}\fi

\bibitem[{Andersen et~al.(2023)Andersen, Zehner, and Goldhammer}]{andersen2023semi}
Nico Andersen, Fabian Zehner, and Frank Goldhammer. 2023.
\newblock Semi-automatic coding of open-ended text responses in large-scale assessments.
\newblock \emph{Journal of Computer Assisted Learning}, 39(3):841--854.

\bibitem[{Ariely et~al.(2022)Ariely, Nazaretsky, and Alexandron}]{ariely2022personalized}
Moriah Ariely, Tanya Nazaretsky, and Giora Alexandron. 2022.
\newblock Personalized automated formative feedback can support students in generating causal explanations in biology.
\newblock In \emph{Proceedings of the 16th International Conference of the Learning Sciences-ICLS 2022, pp. 953-956}. International Society of the Learning Sciences.

\bibitem[{Ariely et~al.(2023)Ariely, Nazaretsky, and Alexandron}]{ariely2023machine}
Moriah Ariely, Tanya Nazaretsky, and Giora Alexandron. 2023.
\newblock {Machine learning and Hebrew NLP for automated assessment of open-ended questions in biology}.
\newblock \emph{International Journal of Artificial Intelligence in Education}, 33(1):1--34.

\bibitem[{Ariely et~al.(2024)Ariely, Nazaretsky, and Alexandron}]{ariely2024causal}
Moriah Ariely, Tanya Nazaretsky, and Giora Alexandron. 2024.
\newblock Causal-mechanical explanations in biology: Applying automated assessment for personalized learning in the science classroom.
\newblock \emph{Journal of Research in Science Teaching}.

\bibitem[{Baffour et~al.(2023)Baffour, Saxberg, and Crossley}]{baffour-etal-2023-analyzing}
Perpetual Baffour, Tor Saxberg, and Scott Crossley. 2023.
\newblock Analyzing bias in large language model solutions for assisted writing feedback tools: Lessons from the feedback prize competition series.
\newblock In \emph{Proceedings of the 18th Workshop on Innovative Use of NLP for Building Educational Applications (BEA 2023)}, pages 242--246.

\bibitem[{Bender et~al.(2021)Bender, Gebru, McMillan-Major, and Shmitchell}]{bender2021dangers}
Emily~M Bender, Timnit Gebru, Angelina McMillan-Major, and Shmargaret Shmitchell. 2021.
\newblock On the dangers of stochastic parrots: Can language models be too big?
\newblock In \emph{Proceedings of the 2021 ACM conference on Fairness, Accountability, and Transparency}, pages 610--623.

\bibitem[{Bovo et~al.(2013)Bovo, Sanchez, H{\'e}guy, and Duthen}]{bovo2013clustering}
Angela Bovo, St{\'e}phane Sanchez, Olivier H{\'e}guy, and Yves Duthen. 2013.
\newblock Clustering moodle data as a tool for profiling students.
\newblock In \emph{2013 Second international conference on E-Learning and E-Technologies in education (ICEEE)}, pages 121--126. IEEE.

\bibitem[{Caliskan et~al.(2017)Caliskan, Bryson, and Narayanan}]{caliskan2017semantics}
Aylin Caliskan, Joanna~J Bryson, and Arvind Narayanan. 2017.
\newblock Semantics derived automatically from language corpora contain human-like biases.
\newblock \emph{Science}, 356(6334):183--186.

\bibitem[{Cao et~al.(2022)Cao, Pruksachatkun, Chang, Gupta, Kumar, Dhamala, and Galstyan}]{cao-etal-2022-intrinsic}
Yang~Trista Cao, Yada Pruksachatkun, Kai-Wei Chang, Rahul Gupta, Varun Kumar, Jwala Dhamala, and Aram Galstyan. 2022.
\newblock On the intrinsic and extrinsic fairness evaluation metrics for contextualized language representations.
\newblock In \emph{Proceedings of the ACL}, pages 561--570.

\bibitem[{Devlin et~al.(2018)Devlin, Chang, Lee, and Toutanova}]{devlin2018bert}
Jacob Devlin, Ming-Wei Chang, Kenton Lee, and Kristina Toutanova. 2018.
\newblock Bert: Pre-training of deep bidirectional transformers for language understanding.
\newblock \emph{arXiv:1810.04805}.

\bibitem[{Fellows(1994)}]{fellows1994window}
Nancy~J Fellows. 1994.
\newblock A window into thinking: Using student writing to understand conceptual change in science learning.
\newblock \emph{Journal of Research in Science Teaching}, 31(9):985--1001.

\bibitem[{Gerard and Linn(2016)}]{gerard2016using}
Libby~F Gerard and Marcia~C Linn. 2016.
\newblock Using automated scores of student essays to support teacher guidance in classroom inquiry.
\newblock \emph{Journal of Science Teacher Education}, 27(1):111--129.

\bibitem[{Goldfarb-Tarrant et~al.(2021)Goldfarb-Tarrant, Marchant, Mu{\~n}oz~S{\'a}nchez, Pandya, and Lopez}]{goldfarb-tarrant-etal-2021-intrinsic}
Seraphina Goldfarb-Tarrant, Rebecca Marchant, Ricardo Mu{\~n}oz~S{\'a}nchez, Mugdha Pandya, and Adam Lopez. 2021.
\newblock Intrinsic bias metrics do not correlate with application bias.
\newblock In \emph{Proceedings of the ACL}, pages 1926--1940.

\bibitem[{Goldfarb-Tarrant et~al.(2023)Goldfarb-Tarrant, Ungless, Balkir, and Blodgett}]{goldfarb-tarrant-etal-2023-prompt}
Seraphina Goldfarb-Tarrant, Eddie Ungless, Esma Balkir, and Su~Lin Blodgett. 2023.
\newblock This prompt is measuring {\textless}mask{\textgreater}: evaluating bias evaluation in language models.
\newblock In \emph{Findings of the ACL}, pages 2209--2225.

\bibitem[{Hattie and Timperley(2007)}]{hattie2007power}
John Hattie and Helen Timperley. 2007.
\newblock The power of feedback.
\newblock \emph{Review of Educational Research}, 77(1):81--112.

\bibitem[{Hubert and Arabie(1985)}]{hubert1985comparing}
Lawrence Hubert and Phipps Arabie. 1985.
\newblock Comparing partitions.
\newblock \emph{Journal of Classification}, pages 193--218.

\bibitem[{Kaneko et~al.(2022)Kaneko, Bollegala, and Okazaki}]{kaneko-etal-2022-debiasing}
Masahiro Kaneko, Danushka Bollegala, and Naoaki Okazaki. 2022.
\newblock Debiasing isn{'}t enough! {--} on the effectiveness of debiasing {MLM}s and their social biases in downstream tasks.
\newblock In \emph{Proceedings of COLING}, pages 1299--1310.

\bibitem[{Kumar et~al.(2019)Kumar, Aggarwal, Mahata, Shah, Kumaraguru, and Zimmermann}]{kumar2019get}
Yaman Kumar, Swati Aggarwal, Debanjan Mahata, Rajiv~Ratn Shah, Ponnurangam Kumaraguru, and Roger Zimmermann. 2019.
\newblock Get {IT} scored using {AutoSAS} — an automated system for scoring short answers.
\newblock In \emph{Proceedings of AAAI}, pages 9662--9669.

\bibitem[{Le~Quy et~al.(2023)Le~Quy, Friege, and Ntoutsi}]{le2023review}
Tai Le~Quy, Gunnar Friege, and Eirini Ntoutsi. 2023.
\newblock A review of clustering models in educational data science toward fairness-aware learning.
\newblock In Alejandro Pe{\~{n}}a-Ayala, editor, \emph{Educational Data Science: Essentials, Approaches, and Tendencies: Proactive Education based on Empirical Big Data Evidence}, pages 43--94. Springer Nature Singapore, Singapore.

\bibitem[{Li et~al.(2023)Li, Reigh, He, and Adah~Miller}]{li2023can}
Tingting Li, Emily Reigh, Peng He, and Emily Adah~Miller. 2023.
\newblock Can we and should we use artificial intelligence for formative assessment in science?
\newblock \emph{Journal of Research in Science Teaching}, 60(6):1385--1389.

\bibitem[{Li et~al.(2021)Li, Tomar, and Passonneau}]{li-etal-2021-semantic}
Zhaohui Li, Yajur Tomar, and Rebecca~J. Passonneau. 2021.
\newblock A semantic feature-wise transformation relation network for automatic short answer grading.
\newblock In \emph{Proceedings of EMNLP}, pages 6030--6040.

\bibitem[{Lloyd(1982)}]{lloyd1982least}
Stuart Lloyd. 1982.
\newblock Least squares quantization in {PCM}.
\newblock \emph{IEEE Transactions on Information Theory}, 28(2):129--137.

\bibitem[{Manning(1999)}]{Manning1999found}
H.~Manning, C. D. {\,}~Schutze. 1999.
\newblock \emph{Foundations of statistical natural language processing}.
\newblock MIT Press.

\bibitem[{Martin et~al.(2023)Martin, Kranz, Wulff, and Graulich}]{martin2023exploring}
Paul~P Martin, David Kranz, Peter Wulff, and Nicole Graulich. 2023.
\newblock Exploring new depths: Applying machine learning for the analysis of student argumentation in chemistry.
\newblock \emph{Journal of Research in Science Teaching}.

\bibitem[{Masala et~al.(2021)Masala, Ruseti, Dascalu, and Dobre}]{masala2021extracting}
Mihai Masala, Stefan Ruseti, Mihai Dascalu, and Ciprian Dobre. 2021.
\newblock Extracting and clustering main ideas from student feedback using language models.
\newblock In \emph{Proceedings of AIED}, pages 282--292. Springer.

\bibitem[{McInnes et~al.(2017)McInnes, Healy, and Astels}]{McInnes2017}
Leland McInnes, John Healy, and Steve Astels. 2017.
\newblock hdbscan: Hierarchical density based clustering.
\newblock \emph{The Journal of Open Source Software}, 2(11).

\bibitem[{Mizumoto et~al.(2019)Mizumoto, Ouchi, Isobe, Reisert, Nagata, Sekine, and Inui}]{mizumoto-etal-2019-analytic}
Tomoya Mizumoto, Hiroki Ouchi, Yoriko Isobe, Paul Reisert, Ryo Nagata, Satoshi Sekine, and Kentaro Inui. 2019.
\newblock Analytic score prediction and justification identification in automated short answer scoring.
\newblock In \emph{Proceedings of the 14th Workshop on Innovative Use of NLP for Building Educational Applications}, pages 316--325.

\bibitem[{Nehm and Haertig(2012)}]{nehm2012human}
Ross~H Nehm and Hendrik Haertig. 2012.
\newblock Human vs. computer diagnosis of students’ natural selection knowledge: testing the efficacy of text analytic software.
\newblock \emph{Journal of Science Education and Technology}, 21:56--73.

\bibitem[{Paszke et~al.(2019)Paszke, Gross, Massa, Lerer, Bradbury, Chanan, Killeen, Lin, Gimelshein, Antiga et~al.}]{paszke2019pytorch}
Adam Paszke, Sam Gross, Francisco Massa, Adam Lerer, James Bradbury, Gregory Chanan, Trevor Killeen, Zeming Lin, Natalia Gimelshein, Luca Antiga, et~al. 2019.
\newblock Pytorch: An imperative style, high-performance deep learning library.
\newblock \emph{Advances in neural information processing systems}, 32.

\bibitem[{Pedregosa et~al.(2011)Pedregosa, Varoquaux, Gramfort, Michel, Thirion, Grisel, Blondel, Prettenhofer, Weiss, Dubourg, Vanderplas, Passos, Cournapeau, Brucher, Perrot, and Duchesnay}]{scikit-learn}
F.~Pedregosa, G.~Varoquaux, A.~Gramfort, V.~Michel, B.~Thirion, O.~Grisel, M.~Blondel, P.~Prettenhofer, R.~Weiss, V.~Dubourg, J.~Vanderplas, A.~Passos, D.~Cournapeau, M.~Brucher, M.~Perrot, and E.~Duchesnay. 2011.
\newblock Scikit-learn: Machine learning in {P}ython.
\newblock \emph{Journal of Machine Learning Research}, 12:2825--2830.

\bibitem[{Rastrollo-Guerrero et~al.(2020)Rastrollo-Guerrero, G{\'o}mez-Pulido, and Dur{\'a}n-Dom{\'\i}nguez}]{rastrollo2020analyzing}
Juan~L Rastrollo-Guerrero, Juan~A G{\'o}mez-Pulido, and Arturo Dur{\'a}n-Dom{\'\i}nguez. 2020.
\newblock Analyzing and predicting students’ performance by means of machine learning: A review.
\newblock \emph{Applied Sciences}, 10(3):1042.

\bibitem[{Reimers and Gurevych(2019)}]{reimers2019sentence}
Nils Reimers and Iryna Gurevych. 2019.
\newblock {Sentence-BERT: Sentence Embeddings Using Siamese BERT-Networks}.
\newblock In \emph{Proceedings of EMNLP}, pages 3982--3992.

\bibitem[{Riordan et~al.(2020)Riordan, Bichler, Bradford, King~Chen, Wiley, Gerard, and C.~Linn}]{riordan-etal-2020-empirical}
Brian Riordan, Sarah Bichler, Allison Bradford, Jennifer King~Chen, Korah Wiley, Libby Gerard, and Marcia C.~Linn. 2020.
\newblock An empirical investigation of neural methods for content scoring of science explanations.
\newblock In \emph{Proceedings of the 15th Workshop on Innovative Use of NLP for Building Educational Applications}, pages 135--144.

\bibitem[{Salmon(2006)}]{salmon2006four}
Wesley~C Salmon. 2006.
\newblock \emph{Four decades of scientific explanation}.
\newblock University of Pittsburgh press.

\bibitem[{Schleifer et~al.(2023)Schleifer, Klebanov, Ariely, and Alexandron}]{schleifer2023transformer}
Abigail~Gurin Schleifer, Beata~Beigman Klebanov, Moriah Ariely, and Giora Alexandron. 2023.
\newblock {Transformer-based Hebrew NLP models for short answer scoring in biology}.
\newblock In \emph{Proceedings of the 18th Workshop on Innovative Use of NLP for Building Educational Applications}, pages 550--555.

\bibitem[{Seker et~al.(2022)Seker, Bandel, Bareket, Brusilovsky, Greenfeld, and Tsarfaty}]{seker2022alephbert}
Amit Seker, Elron Bandel, Dan Bareket, Idan Brusilovsky, Refael Greenfeld, and Reut Tsarfaty. 2022.
\newblock Alephbert: Language model pre-training and evaluation from sub-word to sentence level.
\newblock In \emph{Proceedings of the ACL}, pages 46--56.

\bibitem[{Su{\'a}rez et~al.(2020)Su{\'a}rez, Romary, and Sagot}]{suarez2020monolingual}
Pedro Javier~Ortiz Su{\'a}rez, Laurent Romary, and Beno{\^\i}t Sagot. 2020.
\newblock A monolingual approach to contextualized word embeddings for mid-resource languages.
\newblock In \emph{Proceedings of the ACL}, pages 1703--1714.

\bibitem[{Sung et~al.(2019)Sung, Dhamecha, and Mukhi}]{sung2019improving}
Chul Sung, Tejas~Indulal Dhamecha, and Nirmal Mukhi. 2019.
\newblock Improving short answer grading using transformer-based pre-training.
\newblock In \emph{Proceedings of AIED}, pages 469--481.

\bibitem[{Tansomboon et~al.(2017)Tansomboon, Gerard, Vitale, and Linn}]{tansomboon2017designing}
Charissa Tansomboon, Libby~F Gerard, Jonathan~M Vitale, and Marcia~C Linn. 2017.
\newblock Designing automated guidance to promote productive revision of science explanations.
\newblock \emph{International Journal of Artificial Intelligence in Education}, 27:729--757.

\bibitem[{Toulmin(2003)}]{toulmin2003uses}
Stephen~E Toulmin. 2003.
\newblock \emph{The uses of argument}.
\newblock Cambridge University Press.

\bibitem[{Vinh et~al.(2009)Vinh, Epps, and Bailey}]{vinh2009information}
Nguyen~Xuan Vinh, Julien Epps, and James Bailey. 2009.
\newblock Information theoretic measures for clusterings comparison: Is a correction for chance necessary?
\newblock In \emph{Proceedings of ICML}, pages 1073--1080.

\bibitem[{Wambsganss et~al.(2023)Wambsganss, Su, Swamy, Neshaei, Rietsche, and K{\"a}ser}]{wambsganss2023unraveling}
Thiemo Wambsganss, Xiaotian Su, Vinitra Swamy, Seyed Neshaei, Roman Rietsche, and Tanja K{\"a}ser. 2023.
\newblock {Unraveling Downstream Gender Bias from Large Language Models: A Study on AI Educational Writing Assistance}.
\newblock In \emph{Findings of the Association for Computational Linguistics: EMNLP 2023}, pages 10275--10288.

\bibitem[{Wulff et~al.(2022)Wulff, Buschh{\"u}ter, Westphal, Mientus, Nowak, and Borowski}]{wulff2022bridging}
Peter Wulff, David Buschh{\"u}ter, Andrea Westphal, Lukas Mientus, Anna Nowak, and Andreas Borowski. 2022.
\newblock Bridging the gap between qualitative and quantitative assessment in science education research with machine learning—a case for pretrained language models-based clustering.
\newblock \emph{Journal of Science Education and Technology}, 31(4):490--513.

\bibitem[{Xiao et~al.(2016)Xiao, Ye, Esteves, and Rong}]{xiao2016using}
Chengwei Xiao, Jiaqi Ye, Rui~M{\'a}ximo Esteves, and Chunming Rong. 2016.
\newblock Using spearman's correlation coefficients for exploratory data analysis on big dataset.
\newblock \emph{Concurrency and Computation: Practice and Experience}, 28(14):3866--3878.

\end{thebibliography}

%


\onecolumn
\appendix
\section*{Appendix 1}
\label{appA}
\begin{table}[H]
\centering
\begin{tabular}{|>{\raggedright\arraybackslash}p{0.3\textwidth}|>{\raggedright\arraybackslash}p{0.65\textwidth}|}
\hline
\textbf{Item} & \textbf{Text} \\
\hline
Smoking item & The smoke from cigarettes contains several harmful substances, including the gas carbon-monoxide (CO). CO is released from cigarettes while smoking, and has a stronger tendency than oxygen to bind to Hemoglobin. Explain how high levels of CO make it difficult for smokers to exercise.\\
\hline
Anemia item & A person was found to have low levels of red blood cells in his blood test (anemia). This person complained to his doctor about weakness and difficulty to exercise. Explain how low levels of red blood cells make it difficult for people with anemia to exercise.\\
\hline
\end{tabular}
\caption{The constructed response items (reproduced from \citet{ariely2024causal}; original responses are in Hebrew). } 
\label{tab:items}
\end{table}

\begin{table}[H]
\centering
\begin{tabular}{|>{\raggedright\arraybackslash}p{0.3\textwidth}|>{\raggedright\arraybackslash}p{0.65\textwidth}|}
\hline
\textbf{Cluster description} & \textbf{Exemplifying response} \\
\hline
Full explanations: All/most of the conceptual components and the underlying causal relations are present.  & “Red blood cells bind oxygen and transfer it in the bloodstream, from the lungs where it is absorbed, to all the cells of the body. A low amount of red blood cells in the body leads to the transfer of less oxygen to the body's cells. Since oxygen is one of the reactants in the process of cellular respiration - the energy production process, less oxygen reaching the cells leads to damage to this process. Thus, less available energy is produced in the body's cells and this impairs their function, which leads to fatigue and difficulty in performing physical activity.” (Anemia item) \\
\hline
Gaps in causal connections: All/most of the conceptual components are present, but all/most of the causal relations are missing. &  “The CO binds to the red blood cell instead of the oxygen and thus oxygen does not reach the cells of the body and then cellular respiration does not occur and the body cannot produce energy and thus it stops physical activity due to lack of energy.” (Smoking item) \\
\hline
Specific sequential stages are missing and causal relations are often missing. &  “CO gas is known to bind to Hemoglobin with a stronger tendency than oxygen. When CO binds to Hemoglobin, it takes the oxygen’s place, so much less oxygen is transported from place to place and enters the cells. Lack of oxygen in the cells leads to less production of ATP molecules. Since energy is required for physical exercise, the result is that the person gets tired quickly and has difficulty exercising.” (Smoking item) \\
\hline
Many sequential stages are missing and causal relations are often missing as well. & “Red blood cells carry the oxygen (because of Hemoglobin). When there is anemia, then there is a low amount of red blood cells and thus a low amount of oxygen reaches the muscles.” (Anemia item)\\
\hline
No explanation: All/most of the sequential stages and the underlying causal relations are missing/irrelevant responses. &  “I don’t know” \newline “Anemic people are tired because they have few red blood cells.” (Anemia item) \\
\hline
\end{tabular}
\caption{Examples of student responses and their classification into the KPs that were derived from the expert scoring according to the theory-driven rubric (reproduced from \citet{ariely2024causal}; original responses are in Hebrew). } 
\label{tab:examples}
\end{table}

\end{document}